\documentclass[runningheads,a4pasper]{llncs}

\usepackage{amssymb}
\usepackage{graphicx}

\usepackage{url}
\urldef{\mailsa}\path|{dmargan, amestrovic, smarti}@uniri.hr|

\begin{document}

\mainmatter

\title{Multilayer Network of Language: a Unified Framework for Structural Analysis of Linguistic Subsystems}

\titlerunning{Multilayer Network of Language}

\author{Domagoj Margan, Ana Me\v{s}trovi\'c, Sanda Martin\v{c}i\'c-Ip\v{s}i\'c}

\authorrunning{Domagoj Margan, Ana Me\v{s}trovi\'c, Sanda Martin\v{c}i\'c-Ip\v{s}i\'c}

\institute{Department of Informatics,\\
University of Rijeka,\\
Radmile Matej\v{c}i\'c 2, 51000 Rijeka, Croatia \\
\mailsa\\
}

\maketitle

\section*{Abstract}
Recently, the focus of complex networks’ research has shifted from the analysis of isolated properties of a system toward a more realistic modeling of multiple phenomena - multilayer networks. Motivated by the prosperity of multilayer approach in social, transport or trade systems, we propose the introduction of multilayer networks for language. The multilayer network of language is a unified framework for modeling linguistic subsystems and their structural properties enabling the exploration of their mutual interactions. Various aspects of natural language systems can be represented as complex networks, whose vertices depict linguistic units, while links model their relations. The multilayer network of language is defined by three aspects: the network construction principle, the linguistic subsystem and the language of interest. More precisely, we construct a word-level (syntax, co-occurrence and its shuffled counterpart) and a subword level (syllables and graphemes) network layers, from five variations of original text (in the modeled language). The analysis and comparison of layers at the word and subword levels is employed in order to determine the mechanism of the structural influences between linguistic units and subsystems. The obtained results suggest that there are substantial differences between the networks' structures of different language subsystems, which are hidden during the exploration of an isolated layer. The word-level layers share structural properties regardless of the language (e.g. Croatian or English), while the syllabic subword level expresses more language dependent structural properties. The preserved weighted overlap quantifies the similarity of word-level layers in weighted and directed networks. Moreover, the analysis of motifs reveals a close topological structure of the syntactic and syllabic layers for both languages. The findings corroborate that the multilayer network framework is a powerful, consistent and systematic approach to model several linguistic subsystems simultaneously and hence to provide a more unified view on language.

\section*{Introduction}
Recently, the field of complex networks has shifted from the analysis of isolated network (capturing and modeling one aspect of the examined system) toward the analysis of the family of complex networks simultaneously modeling different phenomena (aspects) of the examined system, or modeling interactions and relationships among different subsystems. The rise of this more realistic framework for a complex network analysis considers different layers, levels or hierarchies for different aspects of the system. In other words, multiple phenomena are characterized by multiple types of links across various levels of representations or various dimensions of relations for multiple subsystems. The multilayer network approach has been addressed in the analysis of real international trade analysis~\cite{Barigozzi2010}, social interactions in the massive online game~\cite{Szell2010}, web-search queries~\cite{Berlingero2011}, in transport and infrastructure~\cite{Kurant2006,Morris2012,Estrada2013,Cardillo2013} and in the examination of the brain's function~\cite{bullmoreEd}. There are variations in formal representation of the multilayer networks~\cite{Kurant2006,DeDomenico2013}, multidimensional networks~\cite{Berlingero2011}, multiplex networks~\cite{Estrada2013,Cardillo2013,Menichetti2013}, interdependent networks~\cite{Gao2012,Morris2012} and networks of networks~\cite{Gao2014,Bianconi2014}. A thorough discussion that compares, contrasts, and translates between notions of multilayer, multiplex, interdependent networks and networks of networks is in~\cite{Kivela2014}, which together with~\cite{Boccaletti2014} presents an detailed overview of multilayer network theory. 

Viewed as a unique, biologically-based human faculty~\cite{Chomsky65}, language has been recognized as the reflection of the human cognitive capacities, both in terms of its structure and its computational characteristics~\cite{Jackendoff2011}. Studying languages at intra- and cross-linguistic levels is of paramount importance in relation to our biological, cultural, historical and social beings~\cite{Baronchelli2012}. Hence, human languages, besides still being our main tools of communication, reflect our history and culture. Language can be seen as a complex adaptive system~\cite{Berkner2009,brighton2005}, evolving in parallel with our society~\cite{steels2011}. 

Various aspects of natural language systems can be represented as complex networks, whose vertices depict linguistic units, while links model their morphosyntactic, semantic, pragmatic, etc. interactions. Thus the language network can be constructed at various linguistic levels: syntactic, semantic, phonetic, syllabic, etc. So far there have been efforts to model the phenomena of various language subsystems and examine their unique function through complex networks. Still, the present endeavors in linguistic network research focus on isolated linguistic subsystems lacking to explain (or even explore) the mechanism of their mutual interaction, interplay or inheritance. Obtaining such findings is critical for deepening our understanding of conceptual universalities in natural languages, especially to shed light on the cognitive representation of the language in the human brain~\cite{hickok2009}.

Therefore, one of the main open questions in linguistic networks is explaining how different language subsystems mutually interact~\cite{Fenk2006,brighton2005}. The complexity of any natural language is contained in the interplay among several language levels. Below the word-level, it is possible to explore the type of phonology, morphology and syllabic subsystem complexity. For example, the phonology subsystem complexity is reflected in the morphology subsystem complexity. On the word-level, the morphology subsystem complexity reflects in the complexity of the word order, syntactic rules and the ambiguity of lexis. Since the word order can be considered as the primary factor (but not the only one) that determines linguistic structure, it is important to explore the subsystems’ interactions by which it is influenced. 

In this research we use the multilayer network framework to explore the structural properties of various language subsystems and their mutual interactions. The multilayer network of a language is constructed for the word (co-occurrence, syntax and shuffled) and subword (syllables and graphemes) language levels. The systematic exploration of layers properties is presented for the Indo-European family of languages: one representative of the Slavic group - Croatian, and one representative of the Germanic group - English. The analysis and comparison of layers is employed in order to determine structural influences and trade-offs between the subsystems of language. 

Our work contributes mainly to the field of linguistic network research by proposing the multilayer network model for language. The multilayer language network model is established on three aspects: the network construction principle, the linguistic subsystem and the language of interest. Moreover, we introduce  the preserved weighted overlap as the measure of word-level layers similarity in weighted and directed networks. Finally, we propose the characterization of word vs. subword layers relationships by correlations of triad significance profiles, as a possible quantification of the inter layer relationships.

\subsection*{Related Work}
\paragraph{The Language Networks.} The pioneering work of Dorogovtsev and Mendes~\cite{Dorogovtsev2001} describes language as a self–organizing network of linked words. The observed word web structure distributions naturally emerge from the evolutionary dynamics.
Masucci and Rodgers~\cite{Masucci2006} investigate the topology of Orwell’s 1984 within the framework of complex network theory. They exhibit local preferential attachment as growth mechanisms of written language and the allocation of a set of preselected vertices that have a structural rather than a functional purpose.
Choudhury and Mukherjee in~\cite{Choudhury2009} provide a suitable framework to model a language from three different perspectives microscopic (utterances), macroscopic (grammar rules and a vocabulary) and mesoscopic (linguistic entities - letters, words or phrases). The authors mainly present an overview of the structure and dynamics at the mesoscopic level. 
Sole et al.~\cite{Sole2010} review the state-of-the-art on language networks and their potential relevance to cognitive science. They also consider the intertwining of language levels related to multiple layers of complexity in terms of the networks of connected words in order to shed light onto the relevant questions concerning language organization and its evolution.
In~\cite{Haitao2014} Cong and Liu provide an extensive insight into the language networks which positions  human language as a multi-level system in the discipline of complex network analysis. Relationships between the system-level complexity of human language (determined by the topology of linguistic networks) and microscopic linguistic features (as the traditional concern of linguistics) are positioned within a holistic quantitative approach for linguistic inquiry, which contributes to the understanding of human language at different granularities. 

\paragraph{The Word-level Networks: Co-occurrence vs. Shuffled.}
The construction of language networks relies on the well-established principles of modeling word interaction from the word order in a sentence or in short from their co-occurrence in text. A substantial part of reported research on language networks is dedicated to a detailed structural analysis of co-occurrence networks interpreting their topological properties in the linguistic context~\cite{Dorogovtsev2001,Masucci2006,Choudhury2009,ITIS2013,Liu2013}. Thus, in the linguistic co-occurrence networks properties are derived directly from the word order in texts by connecting words within a window of certain size or sentence. Still, the open question is how the word order itself is reflected in topological properties of the linguistic network. One approach to address this question is to compare networks constructed from normal texts with the networks from randomized or shuffled texts~\cite{complenet,mipro} and networks constructed from syntax dependencies in texts. 

\paragraph{The Word-level Networks: Syntax.}
The syntactic structure of language is captured through syntax dependency relations between a pair of words in a sentence: the head word – the governor of relationship and the dependent word - the modifier. Syntax dependencies between words are formally expressed by dependency grammar (e.g. a set of productions (rules) in the form of a grammar). The dependency grammar is used to parse the syntactic relationships from a sentence in the form of a syntax dependency tree. Thus, the syntax dependency treebank is the set of syntax dependency trees parsed from the sentences in a corpus.  
Ferrer i Cancho et. al~\cite{Ferrer2004}, in the seed work on syntax complex networks model the syntactic dependency relationships of three languages comparatively (Czech, German, Romanian). The set of analyzed languages is extended to 7 in~\cite{Ferrer2007}, comparing the structure of global syntactic dependency networks.  The results in~\cite{Ferrer2005,Ferrer2007} show that the proportion of syntactically incorrect relationships rises from about 30 \% to a high 50 \% in a co-occurrence networks constructed with a window of size 2 and 3 respectively. 
In~\cite{Liu2008}, based on the comparison of one syntactic dependency network and two co-occurrence networks of Chinese, the authors confirm small-world and scale-free properties, suggesting that scale-free architecture is of essential importance to the syntax subsystem of human language.
Liu et al.~\cite{Liu2011} and Abramov and Meheler ~\cite{Abramov2011} use network parameters derived from the syntax relationships for hierarchical clustering of languages, deriving the model of the genealogical similarity among 15 and 11 languages respectively. The obtained results on syntax networks suggested that a natural approach to modeling human language is considering the structure of the syntactic dependency relationships besides the simple word-order relations reflected in co-occurrence networks. 

\paragraph{The Subword-level Networks: Syllables.}
The coherent results from language networks involving units smaller than words, such as syllables~\cite{Peng2008,Soares2005}, phonemes~\cite{Abersman2010} or morphemes~\cite{Abramov2011} are still missing. Morphological networks for English and German are presented in~\cite{Abramov2011} and the network properties are expressed in terms of graph entropy measures.
So far, syllable networks have been constructed exclusively for Portuguese~\cite{Soares2005} and Chinese~\cite{Peng2008}. Syllables are a natural intermediate level in the analysis of spoken (as opposed to written) language, since they carry prosody during pronunciation~\cite{NacinovicInFuture2011}. The investigation of syllables is particularly interesting for their role in language acquisition. Children begin to learn language through syllables, culminating in the development of their mental lexicons~\cite{hickok2009,Feldman2010}. The model of language acquisition was recreated with humanoid robots using syllables as basic units~\cite{Lyon2012} or by artificial agents~\cite{Oudeyer2001}. Both studies witness the complexity of a language syllable system as an important factor in language acquisition.

\paragraph{The Subword-level Networks: Graphemes.}
Language is written with a set of abstract orthographic symbols (letters of an alphabet) – graphemes. Graphemes are the smallest semantically distinguishing units (the basic linguistic units) in a written language, analogous to the phonemes in spoken language. The complex networks of grapheme subsystem of language have been studied sporadically~\cite{kello2009scale}. Kello and Beltz analyzed the structure of the complex network constructed from the orthographic wordform lexicon, where words are connected if one is a substring of the other. Phonemes have attracted more attention since many psycholinguistics studies regarding the representation of mental lexicon used for speech production, word recognition and language processing have been reported~\cite{vitevitch2008what,gruenenfelder2009lexical,mukherjee2009self,arbesman2010comparative,vitevitch2014insights}. Phonetic networks are typically constructed from words in a lexicon, establishing links among phonetically similar words – differing in one phoneme. 

\paragraph{Network Motifs for Language.}
Motifs are subgraphs defined as simple building blocks of directed complex networks~\cite{Milo2002}. Motifs are used to detect the structural similarities and differences between networks on the local level. In~\cite{Milo2004} the significance profiles of motifs derive several superfamilies of networks - the language networks forms one supra family based on the triad significance profile. Binemann et al. in~\cite{Binemann2012} use motifs to quantify the differences between natural and generated language. The frequencies of three-vertex and four-vertex motifs for six languages are compared with the generated language from n-gram statistical model (n-grams are a sequence of n units from a given text). The authors show that the four-vertex motifs are directly interpretable by semantic relations of polysemy and synonymy. An initial attempt to analyze undirected triads in a multiplex network, by representing positive and negative social interactions of game players in massive online game is reported in~\cite{Szell2010}.

\paragraph{Croatian vs. English.}
A short recapitulation of the main properties of the Croatian and English languages establishes the linguistic framework needed for the comparison across languages as well as for the interpretation of insights into their structural characteristics. Croatian is a highly flective Slavic language and words can have seven different cases for singular and seven for plural, genders and numbers. The Croatian word order is mostly free, especially in non-formal writing. These features place Croatian among morphologically rich and mostly free word-order languages. 
English grammar has minimal inflection compared with most other Indo-European languages, therefore it is considered to be analytic. English word order is almost exclusively subject-verb-object. Both languages are characterized by an accentuation  system developed on syllables. 

English has been studied extensively in a complex networks framework~\cite{Masucci2006,Choudhury2009,Sole2010,Liu2011,Liu2013,Binemann2012}, still no systematic effort explaining the effects of mutual interaction of different subsystems has been reported. So far the Croatian has been quantified in a complex networks framework based on the word co-occurrences~\cite{ITIS2013,mipro,Liu2013} and compared with shuffled counterparts~\cite{complenet,mipro}. The syntax relationships of Croatian as well as syllabic subword units are novelty characterized through the lenses the analysis of complex networks in this research. 
 
\section*{Methods}
More details about complex networks analysis and the definition of measures can be found in~\cite{Newman2010}. Here we list a short definition of measures needed for the exploration of network layers. The network $G = (V,E)$ is a pair of a set of vertices $V$ and a set of links $E$, where $N$ is the number of vertices and $K$ is the number of links. In weighted networks every link connecting two vertices $i$ and $j$ has an associated weight $w_{ij}$ and the number of network components is denoted by $\omega$. 

For every two connected vertices $i$ and $j$ the number of links lying on the shortest path between them is denoted as $d_{ij}$, then the average path length between every two vertices $i,j$ is
$L = \sum_{i,j} \frac{d_{ij}}{N(N-1)}$. If the number of components $\omega > 1$, $L$ is computed for the largest connected component in network.

For weighted networks the clustering coefficient of a vertex $i$ is defined as the geometric average of the subgraph link weights: $c_{i} = \frac{1}{k_{i}(k_{i}-1)} \sum_{j,k} (\hat{w}_{ij} \hat{w}_{ik} \hat{w}_{jk})^{1/3}$, where $k_{i}$ is the degree of the vertex $i$, and the link weights $\hat{w}_{ij}$ are normalized by the maximum weight in the network $\hat{w}_{ij} = w_{ij}/\max(w)$. The value of $c_{i}$ is assigned to 0 if $k_{i} < 2$. The average clustering coefficient of a network is defined as the average value of the clustering coefficients of all vertices in an undirected network: $C = \frac{1}{N}\sum_{i} c_{i}$. 

The transitivity of a network is the fraction of all possible triangles present in the network. Possible triangles are identified by the number of triads (two links with a shared vertex): $T=(3\#triangles)/(\#triads)$. 

The in-degree and out-degree $k_{i}^{in/out}$ of vertex $i$ is defined as the number of its in and out nearest neighbors. The in-strength and the out-strength $s_{i}^{in/out}$ of the vertex $i$ is defined as the number of its incoming and outgoing links, that is: $s_{i}^{in/out} =  \sum_{j}w_{ji/ij}$.

The in- and out- selectivity of the vertex $i$ is then defined as proposed in ~\cite{Masucci2006}:
\begin{equation}
e_{i}^{in/out} = \frac{s_{i}^{in/out}}{k_{i}^{in/out}}.
\end{equation}
The power law distribution is defined as: $P(k) \sim k^{- \gamma}$ where $\gamma$ is the power-law exponent. 

\subsection*{Network Motifs Analysis}
Network motifs are connected and directed subgraphs (of three to up to eight vertices) occurring in complex networks at numbers that are significantly higher than those in randomized networks with the same degree distribution~\cite{Milo2002,Milo2004}. Here, we analyze only triads (all possible directed three-vertex subgraphs) by calculating their frequencies, Z-scores and triad significance profiles (TSP).

The scores $Z_{i}$ for each triad $i$ is calculated using equation:
\begin{equation}
Z_{i}=\frac{N_{i}^{orig}-\langle N_{i}^{rand} \rangle}{\sigma_{i}^{rand}},
\end{equation}
where $N_{i}^{orig}$ is the count of appearances of the triad $i$ in the original network, while $\langle N_{i}^{rand}\rangle$ and $\sigma_{i}^{rand}$ are the average and the standard deviation of the counts of the triad $i$ over a sample of randomly generated networks.

The triad significance profile $TSP$ is the normalized vector of statistical significance scores $Z_i$ for each triad $i$ $TSP_i=\frac{Z_{i}}{\sqrt{\sum_{i}Z_{i}^{2}}}$.

\subsection*{The Multilayer Network}
Since language networks can be viewed through different aspects: different levels (e.g. word-level, subword-level), different construction rules (e.g. co-occurrence, shuffle), different languages, etc. there is a need for a general network model that can capture all these aspects in one single framework. Therefore, we propose an application of general multilayer  networks model introduced by Kivela \textit{et al.} in~\cite{Kivela2014} to the multilayer language networks. 

According to~\cite{Kivela2014}, a  multilayer network can have any number $d$ of aspects defined as a sequence $L= \{ L_{a} \}_{a=1}^d$. There is one set of elementary layers $L_{a}$ for each aspect $a$. It is possible to construct a set of layers in a multilayer network by assembling a set of all of the combinations of elementary layers using a Cartesian product $L_{1} \times ... \times L_{d}$. 

The multilayer network is a quadruplet $M = (V_{M}, E_{M}, V, L)$, where $V_{M} \subseteq V \times L_{1} \times ... \times L_{d}$ that contains only the vertex-layer combinations in which a vertex is present in the corresponding layer, and where $E_{M}$ is a set of pairs of the  possible combinations of vertices and elementary layers, $E_{M} \subseteq V_{M} \times V_{M}$. $V$ is a set of all vertices in all layers. Multiplex is a special case of multilayer network, which satisfies the condition that the set of vertices is shared across layers. Thus, in a multilpex network inter layer connections between different layers have 1:1 or 0:1 cardinality of relationships. 

Furthermore we present equations for the calculation of the overlap between two layers. These equations can be applied only to a multiplex network, when two layers share the same vertices (e.g. in our case it is applicable only to the construction aspects of the word level layers in one language). In the next text we use only  $\alpha$ and $\alpha'$ for the shorter notation of the one layer in the multilayer network. 

Jaccard index for link overlap between two network layers $\alpha$ and $\alpha'$ is:
\begin{equation}
J(E_{\alpha}, E_{\alpha'}) = \frac{\mid E_{\alpha} \cap E_{\alpha'}\mid}{\mid E_{\alpha} \cup E_{\alpha'}\mid}.
\end{equation}

In the same way we can calculate the Jaccard index for weight overlap ($W$).

The preserved weighted ratio on intersected links between network layers $\alpha$ and $\alpha'$ is (modified from total weighted overlap~\cite{Menichetti2013}) is:
\begin{equation}
PW(E_{\alpha}, E_{\alpha'}) = \sum_{i,j} \frac{\min (w_{ij^{\alpha}}, w_{ij^{\alpha'}})}{\max (w_{ij^{\alpha}}, w_{ij^{\alpha'}})}.
\end{equation} 

The preserved weighted overlap (WO) is a normalized preserved weighted ratio:
\begin{equation}
WO(E_{\alpha}, E_{\alpha'}) = \frac{PW(E_{\alpha}, E_{\alpha'})}{\mid E_{\alpha} \cap E_{\alpha'} \mid}.
\end{equation} 

\subsection*{Croatian and English Datasets}
The data sets for multilayer Croatian networks are derived from the HOBS corpus - the first version of the Croatian Dependency Treebank~\cite{hobs}. HOBS is extracted as a part of the Croatian National Corpus~\cite{Tadic2007} and annotated at the analytical layer following the Prague Dependency Treebank formalism adapted to Croatian. The corpus size is currently 3,465 sentences (88,045 tokens). 

The English dataset contains 3,829 sentences (94,084 tokens) from the Penn Treebank corpus~\cite{marcus1993,penn}. The size of the extracted Penn subset is intentionally of the same size as HOBS in order to allow for systematic comparisons across the layers, constructed from comparable corpora of different languages.

Multilayer Croatian (HR) and English (EN) networks are constructed from five variations of HOBS and Penn corpora: three on the word-level (syntax, co-occurrence and it’s shuffled counterpart) and two on the subword level (syllables and graphemes). More clearly, five different realizations of the very same text in one language are used to construct the network layers (all weighted and directed) using five different relationships among the linguistic units: syntax (SIN), co-occurrence (CO), shuffled (SHU), syllables (SYL) and graphemes (GR). 

\subsection*{Language Networks Construction}
The language networks construction principle arises form the vary nature of text (and speech), which is always advancing in an onward direction, hence to use directed and weighted links representing relations among linguistic units\cite{Haitao2014,Masucci2006,ITIS2013}. The co-occurrence relation is established between two adjacent words within a sentence (CO), where the direction of link reflects the words sequencing and weight on the link reflects the frequency of words-pair mutual appearance. 

The syntax relationships among word-pairs are parsed from the HOBS and Penn, as well as the text of the original sentences~\cite{marcus1993}. The sentences' boundaries are preserved, since the syntax dependency is inherent to the sentence (SIN). Thus, the sentence boundaries are considered as linkage delimiters for the co-occurrence layers as well. 

\begin{figure}[h]
\includegraphics[width=5in]{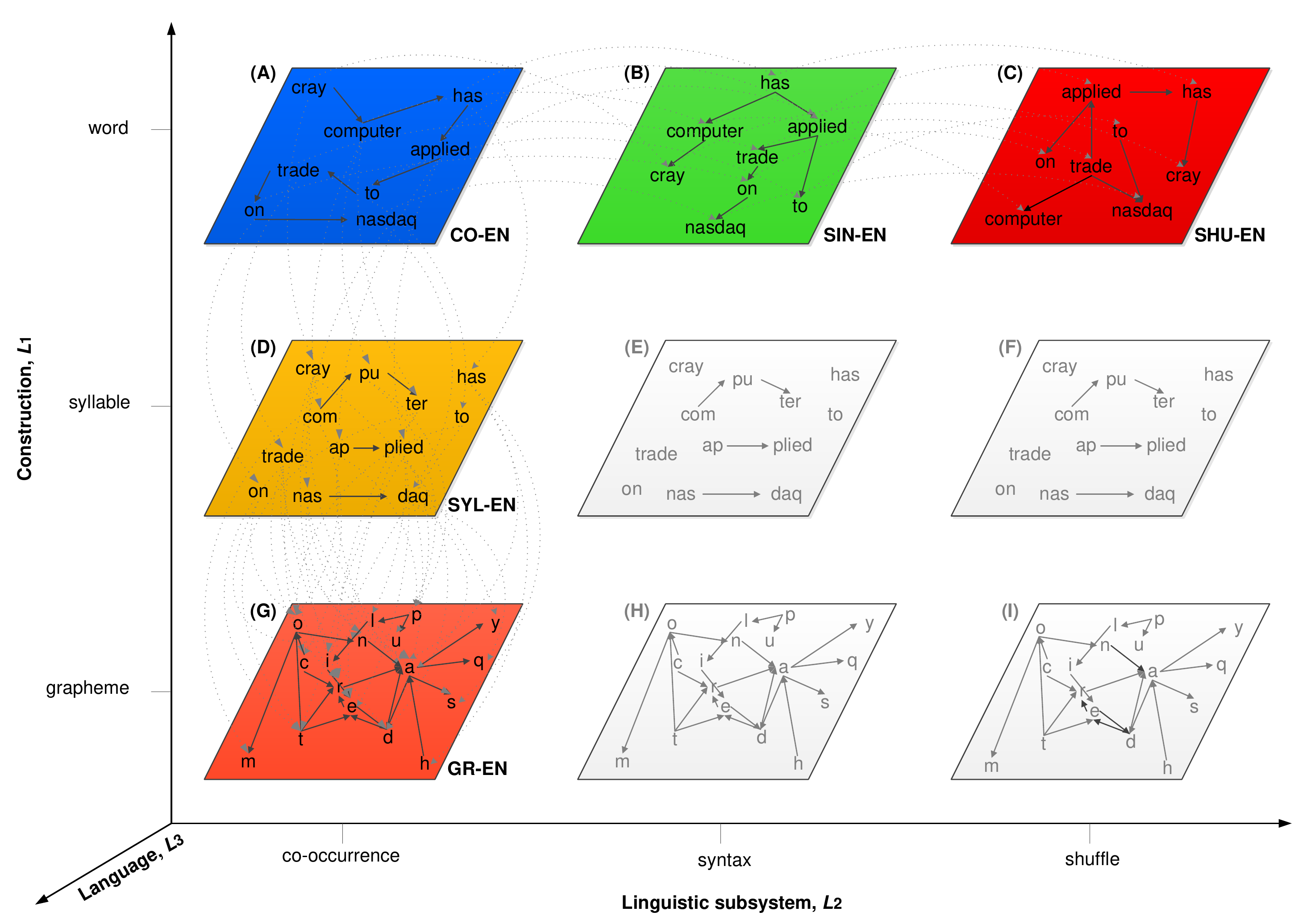} 
\caption{{\bf The multilayer language network.} Three word level layers: (A) co-occurrence; (B) syntax; (C) shuffled; and two subword level layers: syllables (D) and graphemes (G) constructed from the English sentence "\emph{Cray Computer has applied to trade on NASDAQ.}"; according to three aspects of multilayer  network model of language: construction, linguistic subsystem and language. Note- layers (E) and (F); (H) and (I) are gray, since they are disregarded in analysis (identical with layers (D) and (G) respectively).}
\label{fig1}
\end{figure}

Next, the original text is shuffled in order to obtain a shuffled counterpart (SHU), again considering the sentence boundaries. Commonly, the shuffling procedure randomizes the words in the text, transforming the text into a meaningless form. We shuffled the words within the original sentences, preserving the vocabulary size, the word and sentence frequency distributions, the sentence length (the number of words per sentence) and sentence order~\cite{mipro}. Fig.~\ref{fig1} (top part) presents the principles of word-level layers' construction for one sentence. 

Next, we use the Croatian syllabification with a maximal onset algorithm to prepare the last data set – syllables, again from the words in the original sentences. The English syllables are obtained from the dictionary with syllabified words~\cite{Barker}. The process omitted words which were not contained in the syllabified dictionary. The syllable layers are constructed from the co-occurrence of syllables within words (SYL) - presented at the (D) part of Fig.\ref{fig1}. 

Finally, we consider the set of graphemes present in words, where graphemes (GR) represent the most elementary subsystem of each language - orthographical. Since, there are some foreign words present in the used corpora we preserved the original orthographic symbols, resulting in a slightly larger number of graphemes (e.g. in Croatian foreign names contain original diacritic symbols, so we obtained q, w, x, y as Croatian graphemes as well). 

Multilayer language network for this work can be defined with the set $L$ of three aspects: construction $L_{1}$, linguistic subsystem $L_{2}$ and language $L_{3}$, where $L_{1}=$\textit{\{co-occurrence, syntax, shuffle\}}, $L_{2}=\{word, syllable, grapheme\}$ and $L_{3}=\{Croatian, English\} $. 
Therefore, it is possible to have 18 different layers in total ($3x3x2$), although not all the layers are of equal interest. More precisely, one can note that some layers are equal due to the specific construction rules. Since we connect only neighboring syllables within the word, all three layers \textit{(co-occurrence, syllable, Croatian)}, \textit{(syntax, syllable, Croatian)} and \textit{(shuffle, syllable, Croatian)} are equal. The same holds for English syllables, and for graphemes in both languages as well, as shown gray for (E), (F), (H) and (I) parts of Fig.~\ref{fig1}. 

It is worth noticing, that the word-level layers are forming the multiplex networks (have 1:1 inter- connections), while the connections between word and subword layers are not coupled (have N:M inter- connections). 

To sum up, in total we construct ten layers: five of Croatian \textit{(syntax, word, Croatian), (co-occurrence, word, Croatian), (shuffle, word, Croatian), (co-occurrence, syllable, Croatian), (co-occurrence, grapheme, Croatian)}
and five of English language \textit{(syntax, word, English), (co-occurrence, word, English), (shuffle, word, English), (co-occurrence, syllable, English), (co-occurrence, grapheme, English)}, with shortened notations: 
SIN-HR, CO-HR, SHU-HR, SYL-HR, GR-HR, SIN-EN, CO-EN, SHU-EN, SYL-EN and GR-EN. 

Multilayer network construction and analysis was implemented with the Python
programming language using the NetworkX software package developed for the creation, manipulation, and study of the structure, dynamics, and functions of complex networks~\cite{hagberg2008exploring}. The frequencies and triad significant profiles of motifs are obtained with the FANMOD tool~\cite{FANMOD}.

\section*{Results}
Initially we explore the characterization of all isolated layers with the standard set of network measures (see Methods Section). The results for all ten network layers (for both languages) are in Table~\ref{StandardMeasures}. The average path length ($L$) decrease from co-occurrence to syntax, as expected, but interestingly it is of the same range for the syntax and syllabic layer. The clustering coefficient ($C$) (obtained from the undirected versions of the same networks) increases on the syllabic subword level for Croatian and decreased for English. The clustering of English CO and SHU word-levels are higher than their Croatian counterparts. Still, clustering coefficients of SIN layers in both languages are of the same range.

Also, the Croatian syllabic layer has the transitivity higher than the corresponding word layers by one order of magnitude. The numbers of connected components in SYL layers are the highest compared with other layers, and three times higher for English than Croatian. The graphemic layers of both languages exhibit peculiar features due to the small number of vertices, or in other words, due to the high density of GR networks (0.9 - HR;  1.03 - EN).

\begin{table}[!ht]
\begin{center}
\caption{\textbf{The standard network measures for ten layers.}}
\begin{tabular}{|l|l|l|l|l|l|l|l|l|l|l|}
\hline
& \multicolumn{5}{|l|}{\bf CROATIAN} & \multicolumn{5}{|l|}{\bf ENGLISH}\\ \hline
    & CO & SIN & SHU & SYL & GR & CO & SIN & SHU & SYL & GR\\ \hline
$N$ & 23359 & 23359 & 23359 & 2634 & 34 & 10930 & 10930 & 10930 & 2599 & 26\\ \hline
$K$ & 71860 & 70155 & 86214 & 18849 & 491 & 50299 & 52221 & 58920 & 6053 & 333\\ \hline
$L$ & 4.01 & 1.81 & 3.74 & 1.86 & 1.58  & 3.47 & 1.96 & 0.45 & 1.88 & 1.51 \\ \hline
$C$ & 0.167 & 0.120 & 0.182 & 0.255 & 0.636 & 0.286 & 0.153 & 0.295 & 0.057 & 0.838 \\ \hline
$T$ & 0.004 & 0.003 & 0.013 & 0.120 & 0.522 & 0.009 & 0.014 & 0.016 & 0.020 & 0.654 \\ \hline
$\omega$ & 2 & 2 & 2 & 17 & 1  & 3 & 3 & 1 & 54 & 1 \\ \hline
\end{tabular}
\begin{flushleft} Measures ($N$ no. of vertices, $K$ no. of links, $L$ avg. path length, $C$ clust. coeff., $T$ transitivity, $\omega$ no. of components) for co-occurrence (CO), syntax (SIN), shuffled (SHU), syllable (SYL) and grapheme (GR) network layers in Croatian  and English.
\end{flushleft}
\label{StandardMeasures}
\end{center}
\end{table}

\subsection*{Word-level Layers}
\begin{figure}[h]
\includegraphics[width=5in]{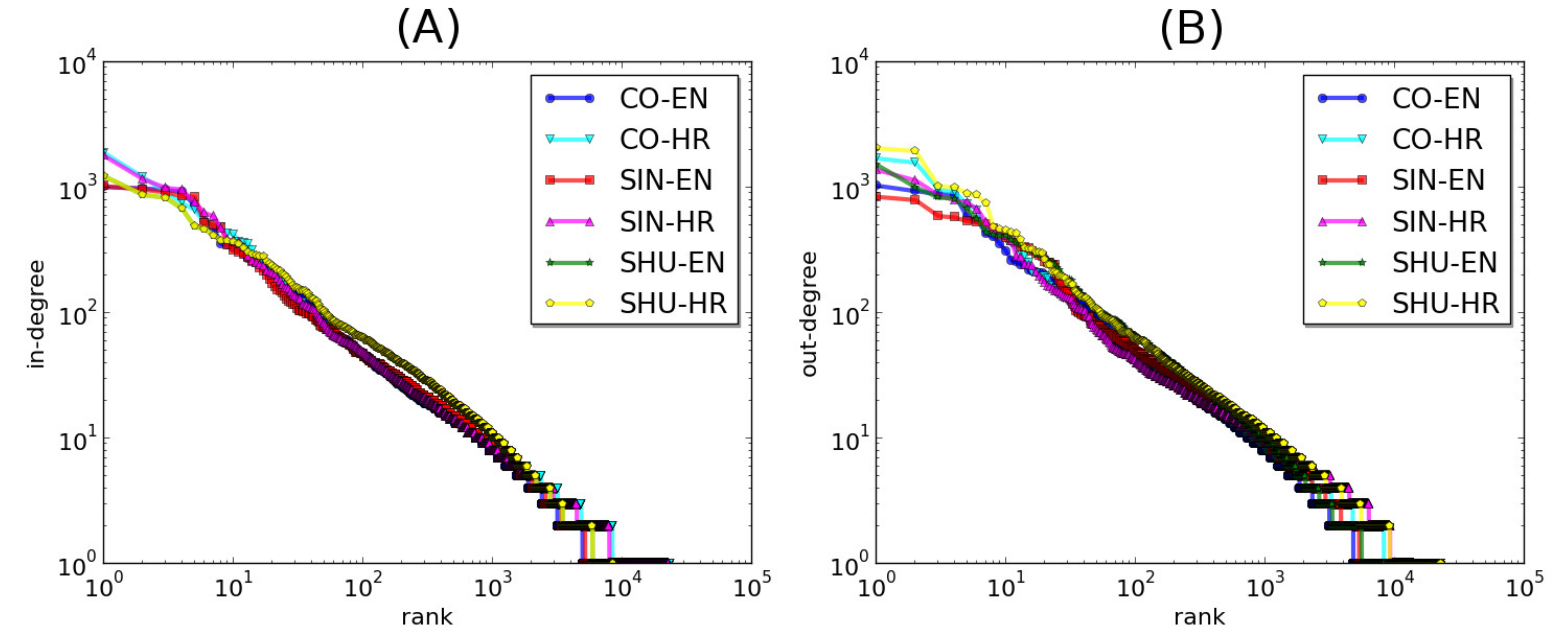}
\caption{\textbf{The word-level layers degree rank distributions.} 
Rank distributions of in- (A) and out- (B) degrees for word level layers: co-occurrence (CO-HR, CO-EN), syntax (SIN-HR, SIN-EN) and shuffled (SHU-HR, SHU-EN).}
\label{fig2}
\end{figure}

For the word-level layers we initially examine the distributions. Fig.~\ref{fig2} shows the rank distributions for in- and out- degrees of word level layers in both languages. The exploitation of the same data source per each language caused the high overlap of exposed distributions. Analogously, the in- and out- strength distributions are overlapped as well, for both languages.  The power-law  $\gamma$ coefficients for all distributions of word-level layers are in a range between 2.14 and 2.49; thus CO, SIN and SHU layers exhibit the power-law distributions for degree and strength regardless of the language.  

\begin{figure}[h]
\includegraphics[width=5in]{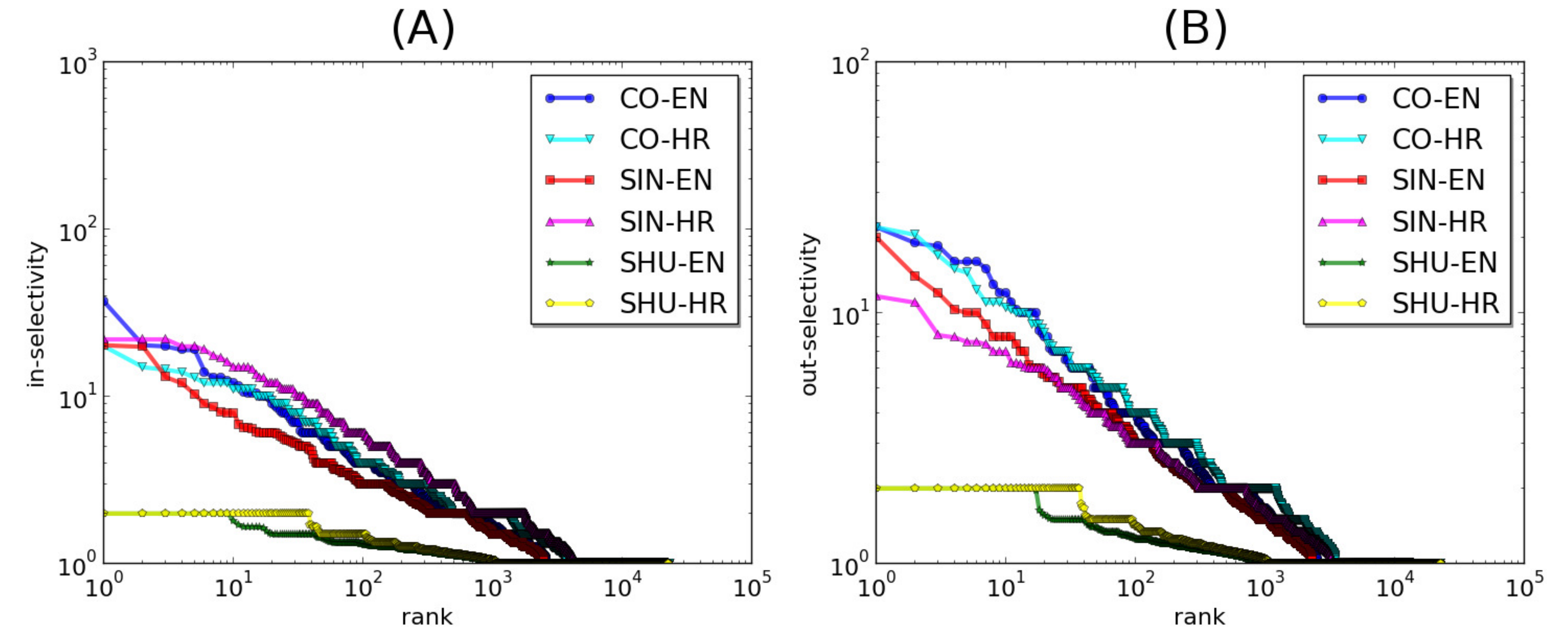} 
\caption{\textbf{The word-level layers selectivity rank distributions.} 
Rank distributions of in- (A) and out- (B) selectivity for word-level network layers: co-occurrence (CO-HR, CO-EN), syntax (SIN-HR, SIN-EN), shuffled (SHU-HR, SHU-EN).}
\label{fig3}
\end{figure}

The potential of selectivity (in- and out-) to differentiate between different text types~\cite{Sisovic2014} or the ability to extract keywords~\cite{BeligaSDSW2014} (identifying and ranking the most representative features of the source text) is restated in this work for the differentiation of language layers as well. Fig.~\ref{fig3} reveals that the rank distributions of in- and out- selectivity for all word-level layers are apart. Selectivity distributions of all three layers co-occurrence (CO), syntax (SIN) and shuffled (SHU) are separated for both languages. 

\begin{figure}[h]
\includegraphics[width=5in]{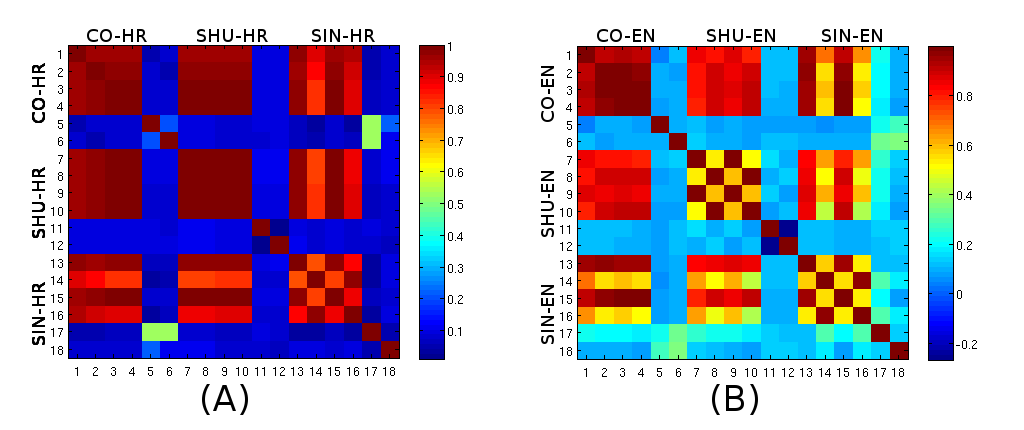} 
\caption{\textbf{Intra and inter layers correlations matrices.}
The correlations matrices for Croatian (A) and English (B): in- \& out- degree, in- \& out- strength and in- \& out- selectivity respectively, presenting inter and intra layer correlations for co-occurrence (CO), shuffled (SHU) and syntax (SIN) word level layers (all p-values $\le$ 0.001). }
\label{fig4}
\end{figure}

The correlation matrices in Fig.~\ref{fig4} show the intra (CO-CO, SHU-SHU and SIN-SIN) and inter layer (CO-SHU, CO-SIN and SHU-SIN) correlations in terms of in- and out- degree, in- and out- strength, in- and out- selectivity distributions. The correlation values for syntax layers of both languages are lower than the corresponding values for the co-occurrence and shuffled layers. Notably, the degree and strength correlation values are higher than the selectivity ones, regardless of the language and layer. Furthermore, Croatian is characterized by higher intra and inter layer correlations than English.

In order to obtain a deeper insight into word-level inter layer relationships we calculated the Jaccard overlap percentage, the percentage of total overlapped weight (W) and the percentage of the preserved weighted overlap (WO) for the overlapping links between word-level layers pairwise (Table ~\ref{tableOvelaps}). The highest percentage of overlapped links is inherent for the intersection of the co-occurrence and syntax layer in both languages, while the overlaps with shuffled layer are expectedly, lower. Furthermore, for both languages the percentage of preserved overlapped weights is relatively high, although slightly lower for English, bearing in mind that less than 20\% of the total possible weights on the total intersected links are preserved.

\begin{table}[!ht]
\caption{\textbf{Overlap of word-level layers.}}
\begin{center}
\begin{tabular}{|l|l|l|l|l|l|l|}
\hline
& \multicolumn{3}{|l|}{\bf CROATIAN} & \multicolumn{3}{|l|}{\bf ENGLISH}\\ \hline
& CO - SIN & CO - SHU & SIN - SHU & CO - SIN & CO - SHU & SIN - SHU \\ \hline
Jaccard & 16.72 \% & 5.47 \% & 4.81 \% & 13.44 \% & 6.31 \% & 5.34 \% \\ \hline
W &  18.96 \% &  6.43 \% & 5.63 \% & 13.58 \% & 6.28 \% & 4.82 \% \\ \hline
WO & 90.6 \% & 76.6 \% & 74.6 \% & 90.00 \% & 74.72 \% & 73.81 \% \\ \hline
\end{tabular}
\begin{flushleft}
The Jaccard overlap percentage, total weighted overlap percentage (W) and preserved weighted overlap percentage (WO) between word-level layers (pairwise) for Croatian and English.
\end{flushleft}
\label{tableOvelaps}
\end{center}
\end{table}

\subsection*{Subword-level vs. Word-level Layers}
Subword-level layers syllabic (SYL) and graphemic (GR) in both languages exhibit the power-law $\gamma$ coefficients between 1.7 and 4.42, which is broader than the observed range of the word-level layers. 

If we compare the syllabic layers of both languages, it is possible to notice some differences between Croatian and English. English syllables are characterized by distributions closer to the word-level layers distributions ($\gamma$ coefficients between 1.87 and 2.14).  The Croatian syllabic layer distributions reveal some deviations ($\gamma$ coefficients are lower - between 1.72 and 1.94). The grapheme layers have $\gamma$ coefficients between 1.7 and 4.16 for Croatian and 2.34 and 4.11 for English. 

\begin{figure}[h]
\includegraphics[width=5in]{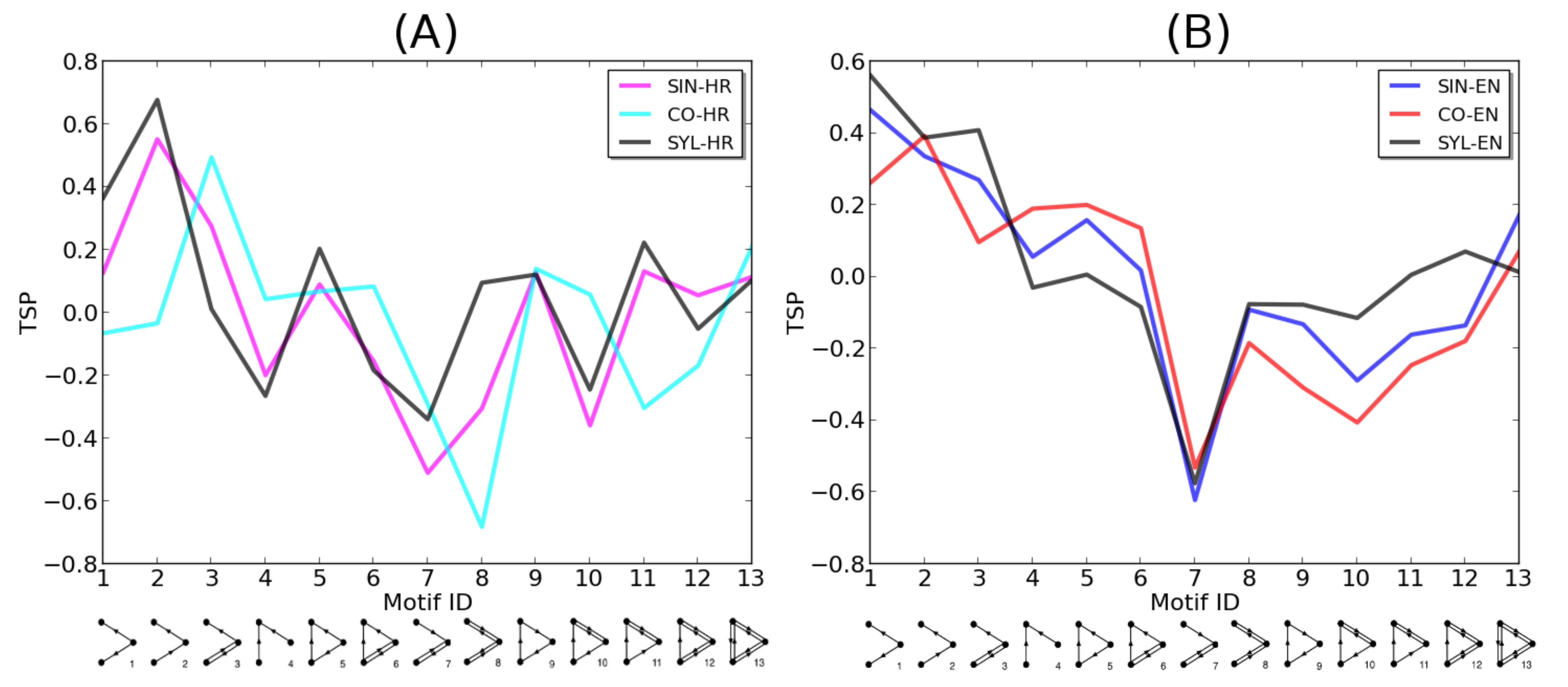} 
\caption{\textbf{The normalized triad significance profiles.}
The normalized triad significance profiles for 13 triadic motifs following the enumeration in~\cite{Milo2004} comparing co-occurrence (CO),  syntax (SIN) and syllables (SYL)layers for Croatian (A) and English (B).}
\label{fig5}
\end{figure}

However, in the multilayer language networks it is interesting to take additional insights of the inter layer relationships, mainly to explore the relationships between word vs. subword layers. For this purpose we introduce the analysis of motifs. We exploited the motif frequencies as well as the normalized triad significance profiles (TSP) of all layers for the analysis. The Pearson correlations for all pairs of network layers in Table ~\ref{MotifiCorr} highlight that motif's frequencies in all layers, with the exception of the graphemic layer are correlated.
Correlations of normalized TSP indicate that SIN and SYL layers in both languages and additionally for English also CO and SIN layers expose similarities. In order to obtain a deeper insight the normalized significance profiles for CO - SIN - SYL layers of Croatian and English per 13 triadic motifs are compared in Fig.~\ref{fig5}. 

\begin{table}[!ht]
\caption{\textbf{The Pearson correlations of triad frequencies and normalized triad significance profiles.}}
\begin{tabular}{ |l|l|l|l|l|l|l|l|l|l|l|}
\hline
\textbf{CROATIAN} & CO-SHU  & CO-SIN & CO-SYL & CO-GR & SHU-SIN & SHU-SYL & SHU-GR & SIN-SYL & SIN-GR & SYL-GR \\ \hline
Freq. & \textbf{0.99} & \textbf{0.95} & \textbf{0.96} & -0.18 & \textbf{0.93} & \textbf{0.96 } & -0.15 & \textbf{0.95 } & -0.20 &  -0.21 \\ \hline
TSP &  0.01  & 0.42  & -0.03 & -0.26 & 0.39  & 0.32  & -0.28 & \textbf{0.83}  & -0.21 & -0.15 \\ \hline
\textbf{ENGLISH} & CO-SHU  & CO-SIN & CO-SYL & CO-GR & SHU-SIN & SHU-SYL & SHU-GR & SIN-SYL & SIN-GR & SYL-GR \\ \hline
Freq. & \textbf{0.93}  & \textbf{0.91}  & \textbf{0.86} & -0.30 & \textbf{0.84 } & 0.74  & -0.17 & \textbf{0.99} & -0.31 & -0.33\\ \hline
TSP  & -0.26 & \textbf{0.92} & 0.73 & 0.39 & 0.04 & 0.35 & -0.27 & \textbf{0.91}& 0.28 & 0.12 \\
\hline 
\end{tabular}
\begin{flushleft} The Pearson correlations of triad frequencies and normalized triad significance profiles (TSP) for all pairs of network layers (co-occurrence (CO), syntax (SIN), shuffled (SHU), syllables (SYL) and graphemes (GR) for Croatian and English (all p-values $\le$ 0.001, emphasized values $\ge$ 0.8).
\end{flushleft}
\label{MotifiCorr}
\end{table}

\section*{Discussion and Conclusion}
The presented findings show that standard network measures on isolated layers exhibit no substantial differences across layers, only slight variations between word and subword levels. Although, if we compare the structural differences across the examined languages there are indications of different principles in their organization. For instance, English is characterized by higher clustering, with the exception of the syllabic layer. The English syllabic layer has 54 components, while Croatian has 17, which is reflected in the low clustering coefficient of English syllables. This is caused by high flectivity of Croatian, where many words share the suffix - the last syllable, which decreases the number of components, and increases the clustering coefficient. This observation raises a question, which properties will the morpheme language subsystem expose during the incorporation into a multilayer language framework?   

Even a standard distribution analysis is not sufficient to take a deeper insight into the mutual influences between subsystems of language. The (in-/out-) degree and  strength distributions of the word-level layers are overlapped due to the same word frequencies reflected from the same data source. Therefore, the standard approach to study the structure of linguistic networks showed no discrepancies among layers. However, the (in-/out-) selectivity values are potentially capable of quantifying differences, namely to show the potential of revealing the interplay among the layers.

The inter layer degree and strength correlations suggest that CO-SHU layers are more related than the CO-SIN, and SIN-SHU pairs, due to the preserving Zipf's law during shuffling~\cite{complenet} (reflecting the utilization of the same data source). In-distributions for syntax layers in both languages have higher values than the corresponding out-distributions, and generally SIN is less inter correlated than the CO and SHU layers. The inter and intra layer correlations in the multilayer language network suggest the manifestation of different governing principles in the syntax structure of the examined languages. The interesting part is that this is the first observable indication of differences between languages manifested in a multilayer analysis framework, which encouraged a deeper investigation. In addition, the selectivity distributions (regardless of side or layer or language) are not correlated, supporting the potential of selectivity as a measure capable to quantify structural differences across language subsystems. Moreover, Croatian exhibits higher correlations then English in general. 

The examination of the word-level layers overlap reveals additional insights into the mutual interplay between the layers. The weighted overlap provides a thorough insight into the intersection of links between network layers. It seems that WO is more appropriate to approximate the overlaps of layers in weighted networks than the commonly employed Jaccard measure. As expected, CO-SIN layers are more overlapped than shuffled pairs, and Croatian syntax is better captured through words co-occurrences than the English. The preserved weights on intersected links indicate that around 10\% of the co-occurrence frequencies are not consistent with overlapped syntax dependencies. The proposed measure of preserved weighted overlap seems adequate to quantify the similarity of word-level layers in weighted and directed multilayer networks of language.    

The subword layer's analysis reveals that the syllabic layer plays an important role in the manifestation of principles governing the construction of word layer, which is different for the examined languages. The graphemic layers, on the other hand, share characteristics, which are reflections of the high density of the graphemic networks (almost complete graphs in both languages). 

The obtained multilayered language analysis results manifest different driving principles beneath the co-occurrence, shuffled, syntactic, syllabic and graphemic layers, which was not obvious through the analysis of isolated layers. 
In order to obtain deeper insight into these relations we utilize the analysis of motifs, which reveal a close topological structure in the syntactic and syllabic layers of both languages. The correlations of the motifs' frequencies are more emphasized in Croatian. The triad significance profiles (TSP) are correlated between syntax and syllables regardless of the language, while English additionally exhibits a correlation between co-occurrence and syntax layers. It seems that the observed TSP correlations reflect the properties of the Croatian - the free word-order which caused different characterizations of the co-occurrence and syntax layers. Moreover, the high flectivity of Croatian is reflected in many suffixes realized by syllables. Therefore, the structure of layers also reflects the morphological properties inherent to the language, which should we examine more deeply in the future. 

Our findings are in line with previous observations in language networks research. For instance, Ferrer i Cancho~\cite{Ferrer2005} reports that the amount of syntactically incorrect links in co-occurrence networks can increase to a high of 70\%, and elaborates: "About 90\% of syntactic relationships take place at a distance lower or equal than two, but word co-occurrence networks lack a linguistically precise definition of link and fail in capturing the characteristic long-distance correlations of words in sentences." This adequately explains the driving principle of the CO-SIN relationships which we have confirmed in this research. Still, an explanation of the linguistic grounding for the SIN-SYL relationships remains an open challenge. 

Our results strongly suggest that there are some properties which are inherent in the word-level layers and not for the subword layers;  while some are inherent in the word-subword relations. More precisely, it seems that syntax and syllables exhibit influences of the same linguistic phenomena.

\paragraph{Conclusion.}
In this research we use the multilayer networks’ framework to explore various language subsystems’ interactions. Multilayer networks are constructed from five variations of the same original text: three on the word-level (syntax, co-occurrence and it’s shuffled counterpart) and two on the subword level (syllables and graphemes). 
The analysis and comparison of layers at word and subword levels is employed in order to determine the mechanism of mutual interactions between different linguistic units.

The presented findings corroborate that the multilayer framework can meet the demands in expressing the complex structure of language. According to these results one can notice substantial differences between the networks' structures of different language layers, which are hidden during the exploration of an isolated layer, regardless of modeled language (e.g. Croatian or English). Therefore, it is important to include all language layers simultaneously in order to capture all language characteristics in the systematic exploration.

The multilayer network framework is a powerful, consistent and systematic approach to model several linguistic subsystems simultaneously and to provide a more general view on language. The word-level layers can be represented as multiplex networks (the coupled links have 1:1 or 0:1 inter-connections), while the  connections between word and subword layers are not coupled (have N:M inter-connections). Hence, defining the unified theoretical model for the multilayer language networks is essential for further endeavors in the research of linguistic networks.

These findings reveal a variety of new and thrilling questions which will open new paths for future research in network linguistics. To conclude, we are at the very beginning of an exciting and challenging pursuit. Hence, our future research plans involve: exploring the relationships of other languages' subsystems (i.e. morphological, phonetic), defining the theoretical model capable of capturing all structural variations of language subsystems' relationships and eventually explain the governing principle of mutual interactions and conceptual universalities in natural languages.





\bibliographystyle{splncs}

\end{document}